# Robust Detection of Intensity Variant Clones in Forged and JPEG Compressed Images

Minati Mishra[1], M. C. Adhikary[2]

*Abstract*

*Digitization of images has made image editing easier. Ease of image editing tempted users and professionals to manipulate digital images leading to digital image forgeries. Today digital image forgery has posed a great threat to the authenticity of the popular digital media, the digital images. A lot of research is going on worldwide to detect image forgery and to separate the forged images from their authentic counterparts. This paper provides a novel intensity invariant detection model (IIDM) for detection of intensity variant clones that is robust against JPEG compression, noise attacks and blurring.*

---

[1]P.G. Department of Information & Communication Technology, F. M. University, Balasore, Odisha, India, minatiminu@yahoo.com
[2]Department of Applied Physics and Ballistics, Fakir Mohan University, Balasore, Odisha, India, mcadhikary@gmail.com

## 1. INTRODUCTION

Cloning is a digital image manipulation technique where a part of an image is copied and pasted into another part of the same image. This is generally done to conceal or recreate an object or a group of objects of interest so as to hide or misrepresent a fact. It is one of the most commonly used image manipulation techniques. Fig. 1 shows two cloned images, where the images on the top are clones of those in the bottom. In the first image, the person on the scene is hidden carefully copy- pasting and blending a part of the scenery whereas in the second image, two more instances of gates are recreated using the same technique.

In a carefully designed cloned image, it is almost impossible to detect the clones visually. Since the cloned region can be of any shape and size and can be located anywhere in the image, it is also not computationally possible to make an exhaustive search of all sizes to all possible image locations. The problem becomes even more critical when the forged image is saved using some lossy compression format such as JPEG or the copied regions have undergone some transformation before/ after being pasted. As a result cloning detection has been remained as a challenging problem in image authentication and digital image forgery detection [1].

In spite of being an important issue for various fields such as forensic investigation, insurance claims etc a little work has been done towards detection of image forgery in general and cloning in special. Out of the handful researches done in this line, most are based on active detection techniques. The active detection methods require pre-embedding of some form of watermark or digital signature to the original image at the time of capture for the forgery to be traced later. Since there are millions of photographs produced every day using an equally high number of cameras available in the market or with people those do not have any watermark inbuilt into, therefore, the active detection methods are too little useful in practical. In contrast to the active methods, the passive methods do not require pre-embedding of any signature or watermark and hence are considered to be more useful in practice.

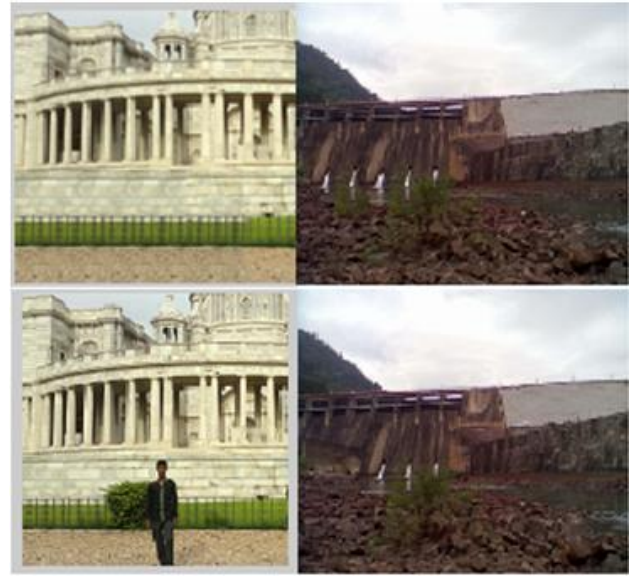

Fig. 1. Images on the top row are clones of those on the bottom row

The most popular passive- blind techniques are based on the principle of overlapped block matching (OLBM). In this method, the test image of size (M x N) pixels is first segmented into (M-b+1) x (N-b+1) overlapping blocks by sliding a window of size (b x b) pixels along the image from top-left corner to bottom-right corner by one pixel at a time. Then the blocks are compared for matches to find the duplicated regions [2]. As this method follows a simple one-to-one pixel intensity matching, the success of this method depends on the fact that the copied region has not undergone any transformation before or after being pasted. The other important concerns of this method are the time complexity and the selection of a proper block size. The time complexity of the algorithm is as high as O (lR2) where, l= b x b, the block size and R = (M-b+1) x (N-b+1) is the total number of overlapped blocks. A larger sized block may reduce the detection accuracy whereas smaller sized blocks generate more false positives hence selecting a proper block size is important.

By representing each block as a vector and then performing a lexicographic sorting on those, duplicated blocks can be arranged into successive rows and can easily be detected by reducing the time complexity to O (lR). Rate of false positives are reduced by measuring the block shift against a predefined threshold [2] [3].Authors in their papers [4] [5] have further reduced the time complexity to O (lTlogT) where, T= (M/2-b+1) *(N/2-b+1) first by decomposing the image into wavelets by DWT and then considering only the LL sub-band.

The simple OLBM and the DWT based methods detect duplicated regions in forged images only when both the original and copied regions have same pixel intensities but fail to detect any forgery if the copied region has undergone some intensity changes or transformations. The DCT and PCA based methods as shown in [1 - 3] provides robustness against intensity changes as well as provides better time complexity but fail when the images have been saved in lossy compression format after the forgery is performed. Kekre et. al [6] in their paper has suggested a hash based tampering detection technique that does not require pre-embedding of watermark-digital signature but requires the original image for comparison, which is not again an effective solution as it is not always possible to get the original image for a comparison.

Luo, Huang and Qiu [7] in their paper suggested a seven feature based duplicate region detection technique that is robust against JPEG compression as well as robust against Additive White Gaussian Noise (AWGN) and blurring but fails when the duplicated regions have different pixel intensities. A comprehensive review of various methods available in literature is provided in [8] but none of the methods provides a solution for tampered images where the clones have different pixel intensities as well as the image is saved in a lossy compression format after the tampering is done. Therefore, in this paper we have presented a robust passive- blind method of forgery detection that effectively detects intensity variant clones in images those are saved in JPEG compressed format after the forgery is done.

The rest of this paper has been organized as follows. Section II describes the proposed algorithm, results and discussions are presented in section III and the paper has been concluded in section IV.

## 2. INTENSITY INVARIANT DETECTION MODEL FOR JPEG COMPRESSED IMAGES (IIDMJPEG)

### A. Selection of Features

Intensity invariant detection can be achieved by applying DCT to the blocks. But since JPEG compression introduces a kind of blocking artifact [9] to the images, simple DCT coefficient matching cannot be used as a solution in such images. Therefore, it is required to select a set of representative characteristic features so that those will be able to detect clones surpassing the intensity variation as well as the blocking artifact. At the same time it should provide minimum redundancy and should be of minimum length so as to provide better time complexity. Keeping all these goals in mind and from our experience with the nature of discrete cosine transformation, we have selected four characteristic features {C1, C2, C3, C4} those can be used to correctly represent a block. Because the Y-channel pixels are weighted averages of the R, G and B channel pixel values so, in our feature vector, we only consider the Y-channel unlike the Weiqi Luo's algorithm [8] that considers all the four channels for feature selection. By carefully studying hundreds of DCT blocks, it has been found that the first coefficient (the left-topmost, (1, 1) coefficient) though can be considered as one of the features but it provides higher redundancy. Hence, we considered the (1, 2) and (2, 1) coefficients as two of our characteristic features (C1 and C2) instead of the (1, 1) coefficient. The 3rd and the 4th features are selected obeying one of the important characteristic of the discrete cosine transformation that lays most of the significant energy in the top-left corner of the block [8]. The average of the upper triangular DCT coefficients with respect to the DCT coefficient block sum therefore, is considered as 3rd feature and the average of top-most 50% rows of the DCT coefficients with respect to the DCT coefficients block sum as the 4th feature. A pictorial representation of the selection procedure is shown in Fig. 2. The first two characters C1 and C2 are selected from the DCT coefficient blocks as shown in Fig. 2 (a). C3= sum of the shaded coefficients as shown in Fig. 2 (b)/ sum of the coefficients of the whole block. C4= sum of the shaded coefficients as shown in Fig. 2(c)/ sum of the coefficients of the whole block. The steps of the proposed algorithm are given in the following section.

### B. The proposed IIDMJPEG Algorithm

**STEP1:** calculate the Y-channel pixel intensities of the test image using the formula:

$$Y = 0.299R + 0.587G + 0.114B \quad (1)$$

**STEP2:** Divide the Y channel image $f_Y$ into overlapping blocks of size $b$ x $b$ and store $YB$.

**STEP3**: For each block $YB_i$, $(i=1, 2 \ldots R = (M-b+1)$ x $(N-b+1))$, compute the DCT coefficients and represent those by $DYB_i$.

**STEP4**: Compute the feature vectors $\vec{U}_i$ $(i=1, 2 \ldots R = (M-b+1)$ x $(N-b+1))$, consisting of four-characteristic features $C_j$, (j =1, 2, 3, 4) as follows:

i. $Ci1 = DYBi\ (2, 1)$.
ii. $Ci2 = DYBi\ (1, 2)$.
iii. $Ci3$ = Sum (Upper triangular coefficients of $DYBi$)/Sum of coefficients of $DYBi$.
iv. $Ci4$ = Sum (coefficients of first 50% rows of $DYBi$)/Sum of coefficients of $DYBi$.

**STEP5**: Arrange the feature vectors $\vec{v_i} = \{Ci1, Ci2, Ci3, Ci4\}$ into an $R \times 4$ matrix $T$ where, $R = (M-b+1) \times (N-b+1)$ are the number of overlapped blocks in a single channel of the image.

**STEP6**: Sort the rows of the matrix $T$ in lexicographic order.

**STEP7**: Compare the neighbouring rows of the sorted matrix to find the duplicate blocks.

**STEP8**: Compute the shift distance $d_i\ (dx, dy)$ for each pair of duplicate blocks and the number of blocks ($Nc$) belonging to two connected regions.

**STEP9**: Declare the duplicate blocks as clones if $d_i >= TH1$, a predefined threshold distance and $Nc >= TH2$, a predefined threshold count.

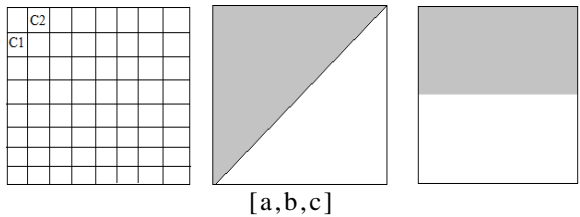

[a,b,c]

Fig. 2. Block DCT Coefficients for Feature Selection

**STEP10**: Process the image with the morphological closing operation for removing the noise, if remained any.

Given below is a brief presentation of the morphological image processing operations, before the results of this proposed method are discussed.

*C. Morphological Image Processing*

Morphological image processing is a collection of non-linear operations related to the shape or morphology of features in an image. It contributes a wide range of operators to image processing, all based around a few simple mathematical concepts from set theory. The operators are particularly useful for the analysis of binary images and common usages include edge detection, noise removal, image enhancement and image segmentation. These operators typically explore an image with a small shape or template known as a structuring element. The structuring element is positioned at all possible locations in an image and is compared with the corresponding neighborhood of pixels to produce a noise free, enhanced resultant image. All most all morphological image processing algorithms typically are based on two primitive operations: dilation and erosion [10].

- *Dilation*

For sets A and B in $Z^2$, the dilation of A by B, denoted as $A \oplus B$ is defined by

$$A \oplus B = \{z \mid (\hat{B})_z \cap A \neq \phi\} \qquad (2)$$

Where, $\hat{B}$ is the reflection of set B, the structuring element in dilatation, defined by (3) and $(A)_z$ is the translation of set A by point $z = (z1, z2)$ defined as in (4)

$$\hat{B} = \{w / w = -b\ for\ b \in B\} \qquad (3)$$
$$(A)_z = \{c / c = a + z\ for\ I \in A\} \qquad (4)$$

- *Erosion*

With A and B are sets in $Z^2$, the erosion of A by B, denoted as $A \ominus B$ is defined by

$$A \ominus B = \{z \mid (B)_z \subset A\} \qquad (5)$$

Dilation operation is generally used to expand an image whereas the erosion is to shrink it.

- *Closing*

Closing a set A by a structuring element B is denoted by $A \bullet B$ is defined as the dilation of A by B followed by the erosion of the result by B. The mathematical form of this operation is given by

$$A \bullet B = (A \oplus B) \ominus B \qquad (6)$$

## 3. RESULTS AND DISCUSSIONS

In this section we present the results of detection of our proposed IIDMJPEG algorithm for JPEG images with different compression factors, for images distorted with AWGN for SNR values between 10 to 40 and Gaussian blurring for 3 x 3 to 12 x 12 filter sizes and for standard deviations varying from .5 to 10. The threshold TH1 is set to |dx - dy|>=10 and HT2=100, as it has been seen from the experiments that, two cloned regions generally contain a minimum 100 blocks (block size = 8 x 8, Image size 128 x 128, total number of overlapped blocks=14641) and the duplicates detected below shift distance 10 generally represent similar blocks belonging to long smooth regions of a natural image and should be discarded.

### A. Detection Robustness against JPEG Compression

The method is tested against 480 JPEG compressed test images (40 different images, each compressed with 12 different quality factors between 100% and 9% where, QF=100% is the best quality image and QF=9% implies the poorer quality image. The results of detection for a randomly selected image are given in Fig. 3. Given in Table 1 and Fig. 4, are the average rates of false positives and accuracies with respect to different JPEG QFs as determined for the 40 test images considered for this study.

### B. Detection Accuracy

The detection accuracy is defined in terms of number of pixels correctly detected with respect to the total number of pixels in the actual cloned region. If R1 and R2 denote the actual duplicated regions and if D1 and D2 represent the clones as detected by the algorithm, then accuracy of detection (ACC) [6] and rate of false positives (FP) are determined using the formulae:

$$ACC = \frac{|R1 \cap D1| + |R2 \cap D2|}{|R1| + |R2|} \quad \text{and}$$

$$FP = \frac{|D1 - R1| + |D2 - R2|}{|R1| + |R2|}$$

TABLE I. AVERAGE DETECTION ACC AND FP OF IIDMJPEG ALGORITHM FOR JPEG COMPRESSED IMAGES WITH DIFFERENT QFs

| JPEG QF (in %) | 9 | 17 | 25 | 33 | 42 | 50 | 58 | 67 | 75 | 83 | 92 | 100 |
|---|---|---|---|---|---|---|---|---|---|---|---|---|
| ACC | 0.1338 | 0.2998 | 0.5225 | 0.3437 | 0.5299 | 0.7023 | 0.6402 | 0.7732 | 0.8269 | 0.8416 | 0.8577 | 0.9220 |
| FP | 0.0105 | 0.0165 | 0.0324 | 0.0180 | 0.0069 | 0.0728 | 0.0296 | 0.0692 | 0.0834 | 0.0664 | 0.0262 | 0 |

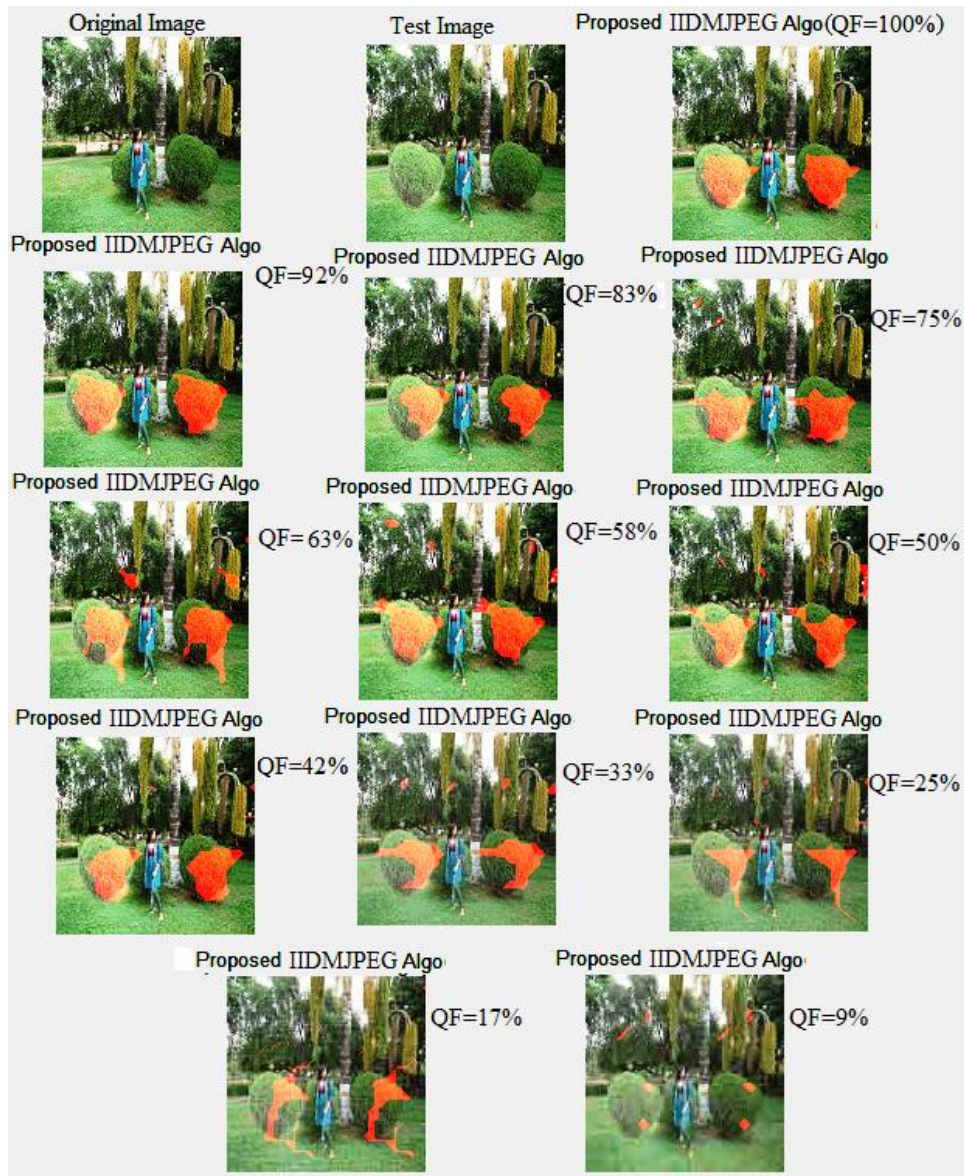

Fig. 3. Detection of intensity variant Clones using the Proposed IIDMJPEG Algorithm in JPEG Compressed Images with Different Quality Factors (Clones detected are shown in Orange)

It can be seen from the results that the detection accuracy of the proposed IIDMJPEG method is much better even after the image is compressed by 67% to 75%. The accuracy reduces beyond this but still the method is able to locate the clones in compressed images with the quality factors as low as 17% that is, with a compression ratio up to 83%.

shown in Fig .5. The average detection accuracy and the rate of false positives of all the 100 images are given in table2 and the corresponding plots are shown in Fig. 6.

### C. Robustness against AWGN

The method is tested against a hundreds of JPEG compressed and noisy images those are contaminated by additive white Gaussian noise (AWGN) with SNR between 5 and 40. The results of detection for one selected image are

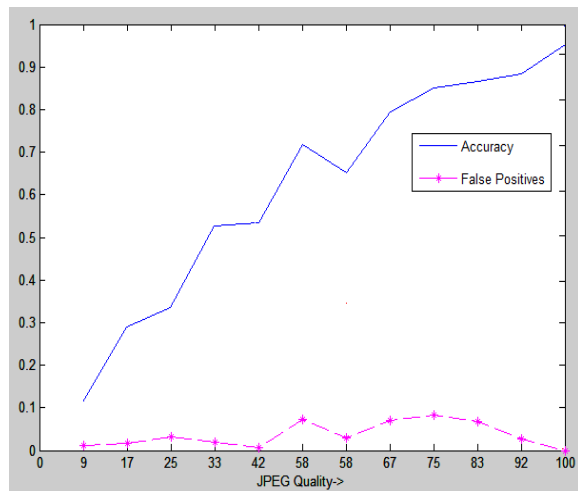

Fig. 4. Quality Factor Vs detection Accuracies and False Positives.

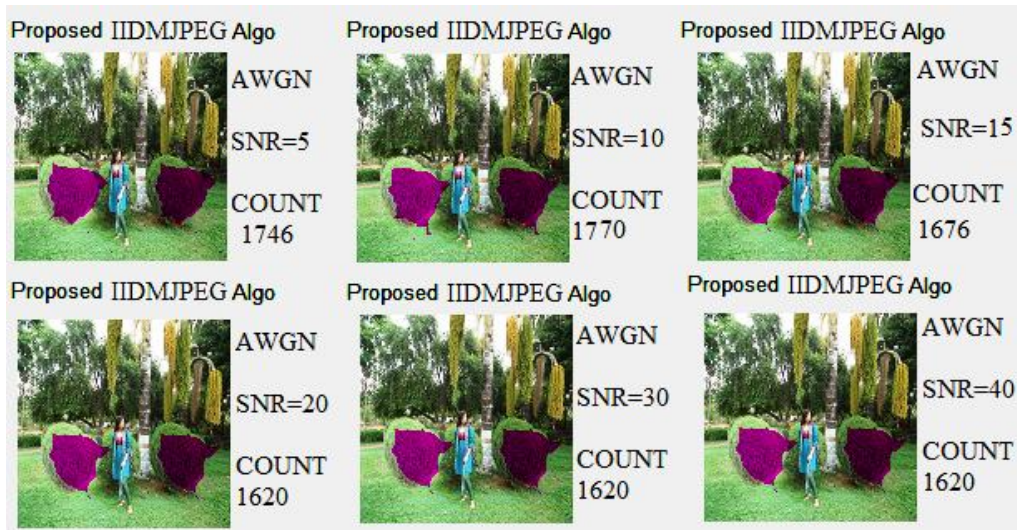

Fig. 5. Detection of intensity variant clones in JPEG Compressed and AWGN distorted images by the proposed IIDMJPEG Algorithm. (Clones detected are shown in purple)

TABLE II. AVERAGE DETECTION ACC AND FP OF THE PROPOSED IIDMJPEG ALGORITHM FOR COMPRESSED AND NOISY IMAGES

| SNR | Detected Cloned pixels | No of False Positives | Positives | ACC | FP |
|---|---|---|---|---|---|
| 5 | 1680 | 220 | 1460 | 0.686 | 0.103 |
| 10 | 1584 | 232 | 1352 | 0.635 | 0.109 |
| 15 | 1604 | 218 | 1386 | 0.651 | 0.102 |
| 20 | 1616 | 242 | 1374 | 0.646 | 0.113 |
| 25 | 1616 | 242 | 1374 | 0.646 | 0.113 |
| 30 | 1616 | 242 | 1374 | 0.646 | 0.113 |

### D. Robustness against JPEG Compressed and Blurred Images

To test the robustness of the algorithm against compression and blurring, all the 100 test images are subjected to Gaussian blurring operation with different values of standard deviation (SD) and variable filter sizes. The results of detection for one selected image are shown in Fig. 7. The average detection accuracy and the rate of false positives for all the 100 images are given in table.3, table.4. Comparisons of Different Block- Based Clone Detection Methods are given in table5. An 'X' in a cell represents: the method does not provide detection against the said category. For example, methods 1 through 9 do not provide detection robustness against intensity variant clones or these methods do not detect intensity variant clones.

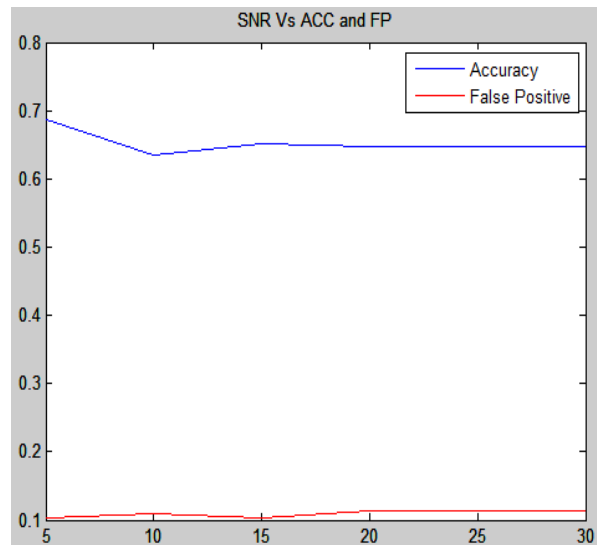

Fig. 6. Average Detection ACC and FP of the proposed IIDMJPEG for compressed and noisy images

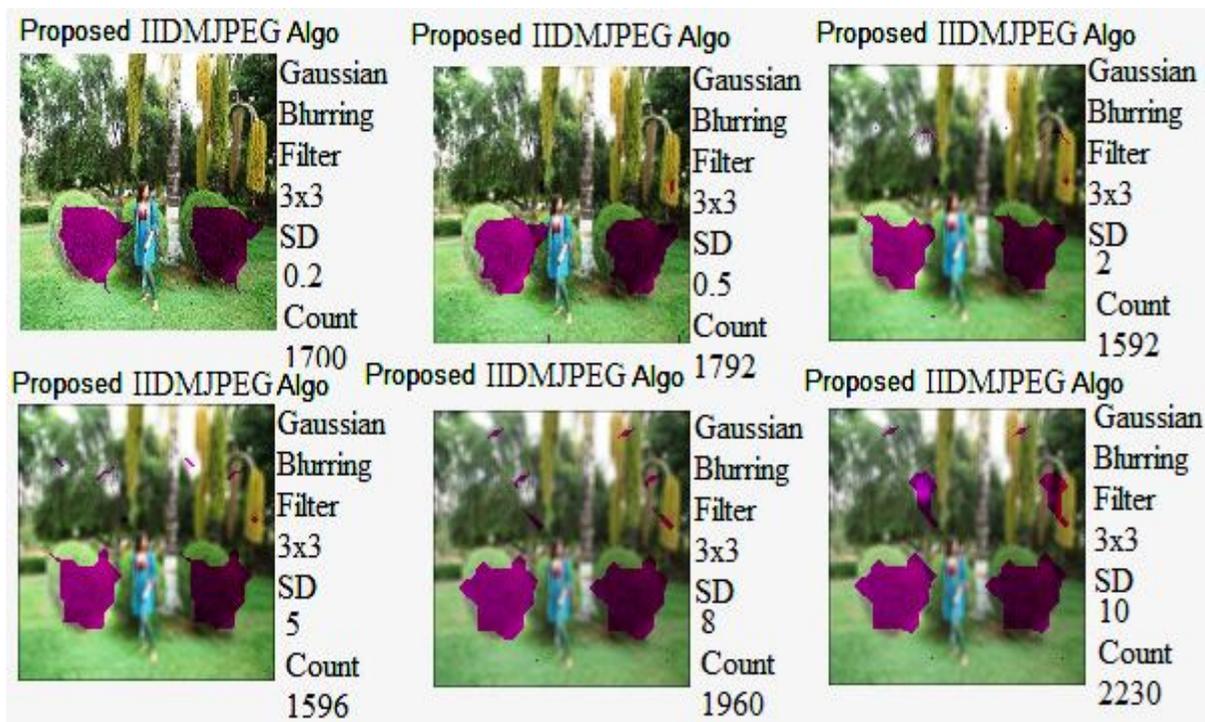

Fig. 7. Detection of intensity variant clones in JPEG Compressed and blurred images by the proposed IIDMJPEG Algorithm. (Clones detected are shown in purple)

TABLE III. AVERAGE DETECTION ACC AND FP OF THE PROPOSED IIDMJPEG ALGORITHM FOR COMPRESSED AND BLURRED IMAGES WITH DIFFERENT FILTER SIZES (SD=0.2)

| Filter Size | Detected Cloned pixels | No of False Positives | Positives | ACC | FP |
|---|---|---|---|---|---|
| 3x3 | 1616 | 242 | 1374 | 0.6463 | 0.1138 |
| 4x4 | 1840 | 656 | 1470 | 0.6914 | 0.3086 |
| 7x7 | 1616 | 242 | 1374 | 0.6463 | 0.1138 |
| 8x8 | 1840 | 656 | 1470 | 0.6914 | 0.3086 |
| 11x11 | 1616 | 242 | 1374 | 0.6463 | 0.1138 |
| 12x12 | 1840 | 656 | 1470 | 0.6914 | 0.3086 |

TABLE IV. AVERAGE DETECTION ACC AND FP OF THE PROPOSED IIDMJPEG FOR COMPRESSED AND BLURRED IMAGES FOR DIFFERENT VALUES OF STANDARD DEVIATION (FILTER SIZE= 3 X 3)

| SD | Detected Cloned pixels | No of False Positives | Positives | ACC | FP |
|---|---|---|---|---|---|
| 0.1 | 1616 | 242 | 1374 | 0.6463 | 0.1138 |
| 0.5 | 1638 | 284 | 1354 | 0.6369 | 0.1336 |
| 2 | 1595 | 222 | 1374 | 0.6463 | 0.1044 |
| 5 | 1642 | 294 | 1348 | 0.6341 | 0.1383 |
| 8 | 1614 | 250 | 1364 | 0.6416 | 0.1176 |
| 10 | 1660 | 264 | 1396 | 0.6566 | 0.1242 |
| 12 | 1660 | 264 | 1396 | 0.6566 | 0.1242 |

## 4. CONCLUSION AND FUTURE SCOPE

Cloning or copy-move forgery is one of the widely used digital image tampering techniques. Detection of these types of forgery are challenging as there exists no incnsistency with respect to lighting and shadowing conditions. The detection is made even difficult through pre and post processing operations such as addition of noise, blurring, lossy compression, intensity changes, geometric transformations etc. In this paper we have proposed an efficienct and robust algorithm that successfully detects clones in tampered images even after the cloned images are subjected to several post processing operations such as lossy compression, noise contamination, blurring, and intensity changes. The work, in future, has to be extended to splicing detection i.e., to detect tampering in images when the forgerd image is a composite of two or more different images.

*Acknowledgment*

We are thankful to Mr. Swetaraj Mohanty, for designing the test images for the experiments and allowing us to use those in this work.

TABLE V. COMPARISON OF DIFFERENT BLOCK- BASED CLONE DETECTION METHODS

| Sl. No. | Method | Image Size/ Block Size(*b*) | No. of Blocks (*R*) | Feature Length (*l*) | Time Complexity | Robustness Against ||||| Remark |
|---|---|---|---|---|---|---|---|---|---|---|---|
| | | | | | | Intensity variant clones (IVC) | JPEG Compression (JC) | IVC + JC | AWG Noise | Gaussian Blur | |
| 1 | Simple OLBM | 128 x 128 \ 8 x 8 | 14641 | 64 | $O(64RlogR)$ | X | X | X | SNR >=13dB | (Filter size, SD)<= (10,10) | Detects cloned regions as small as *b* x *b* |
| 2 | DWT based method | 128 x 128 \ 8 x 8 | 3249 | 64 | $O(64RlogR)$ | X | X | X | SNR >=15dB | (Filter size, SD) <= (5,5) | Clone size should be more than (2*b* x 2*b*). Computation time is minimum |
| 3 | DWTSVDbesed | 128 x 128 \ 8 x 8 | 3249 | 8 | $O(8RlogR)$ | X | X | X | SNR >=15dB | (Filter size, SD) <= (5,5) | Clone size should be more than (2*b* x 2*b*). |
| 4 | IIDMDCT | 128 x 128 \ 8 x 8 | 14641 | 16 | $O(16RlogR)$ | √ | X | X | SNR >=12dB | (Filter size, SD) <= (6,6) | High False +ive rate due to approximation |
| 5 | DCTSVD | 128 x 128 \ 8 x 8 | 14641 | 8 | $O(8RlogR)$ | √ | X | X | SNR >=15dB | (Filter size, SD) <= (5,5) | False +ve rate is high due to approximation. Computation time is high due to real number comparison. |

| 6 | 7 | 8 | 9 | 10 |
|---|---|---|---|---|
| IIDMPCA | DWTPCA | DWTDCT | Weiqi Luo [7] | IIDMJPEG |
| 128 x 128\8 x 8 | 128 x 128\8 x 8 | 128 x 128\8 x 8 | 128 x 128\8 x 8 | 128 x 128\8 x 8 |
| 14641 | 3249 | 3249 | 14641 | 14641 |
| 8 | 8 | 16 | 7 | 4 |
| O(8RlogR) | O(8RlogR) | O(16RlogR) | O(7RlogR) | O(4RlogR) |
| √ | √ | √ | √ | √ |
| X | X | X | JPEG Quality between 58 to 100 | JPEG Quality ranging between 17 to 100 |
| X | X | X | X | √ |
| SNR >=13dB | SNR >=15dB | SNR >=15dB | SNR >=5dB | SNR >=5dB |
| (Filter size, SD) <= (10,10) | (Filter size, SD) <= (5,5) | (Filter size, SD)<=(5,5) | (Filter size, SD)<=(12,10) | (Filter size, SD) <= (12,10) |
| Computation time is small in comparison to DCT and SVD | Computation time is less in comparison to PCA | --- | Rate of accuracy is better for higher QFs | Rate of accuracy is better than JPEGIDM. Detects clones in JPEG images with QF as low as 17. |